\pdfoutput=1
\documentclass[11pt]{article}

\usepackage{ACL2023}

\usepackage{times}
\usepackage{latexsym}

\usepackage[T1]{fontenc}
\usepackage[utf8]{inputenc}

\usepackage{microtype}

\usepackage{inconsolata}

\usepackage{multirow}
\usepackage{booktabs}
\usepackage{graphicx, subfigure}

\newcommand{\standardentail}{\texttt{TE}}
\newcommand{\parallelID}{\texttt{One-to-All}}

\def\@fnsymbol#1{\ensuremath{\ifcase#1\or *\or \dagger\or \ddagger\or
   \mathsection\or \mathparagraph\or \|\or **\or \dagger\dagger
   \or \ddagger\ddagger \else\@ctrerr\fi}}
\newcommand{\ssymbol}[1]{^{\@fnsymbol{#1}}}

\title{All Labels Together: Low-shot Intent Detection with an Efficient Label Semantic Encoding Paradigm}

\author{Jiangshu Du \\ University of Illinois at Chicago \\ \texttt{jdu25@uic.edu} \And Congying Xia \\ Salesforce Research \\  \texttt{c.xia@salesforce.com} \And
        Wenpeng Yin \\ Penn State University \\ \texttt{wenpeng@psu.edu} \AND
        Tingting Liang \\ Hangzhou Dianzi University \\ \texttt{liangtt@hdu.edu.cn} \And
        Philip Yu \\University of Illinois at Chicago \\ \texttt{psyu@uic.edu}}

\begin{document}
\maketitle
\begin{abstract}
In intent detection tasks, leveraging meaningful semantic information from intent labels can be particularly beneficial for few-shot scenarios. 
However, existing few-shot intent detection methods either ignore the intent labels, (e.g. treating intents as indices) or do not fully utilize this information (e.g. only using part of the intent labels).
In this work, we present an end-to-end \parallelID~ system that enables the comparison of an input utterance with all label candidates.
The system can then fully utilize label semantics in this way.
% intent semantic aware model, \parallelID, that fully utilizes label semantics.
Experiments on three few-shot intent detection tasks demonstrate that \parallelID~ is especially effective when the training resource is extremely scarce, achieving state-of-the-art performance in 1-, 3- and 5-shot settings.
Moreover, we present a novel pretraining strategy for our model that utilizes indirect supervision from paraphrasing, enabling zero-shot cross-domain generalization on intent detection tasks. Our code is at \url{https://github.com/jiangshdd/AllLablesTogether}.

\end{abstract}

\section{Introduction}
\label{intro}

Few-shot intent detection aims to identify the intents of user utterances with only a few labeled examples. Recent works can be mainly summarized into three categories:
1) \textbf{Standard classifier-based approaches} \cite{space2, space3, spi, cpft}, which leverage pretrained language models (PLMs) equipped with a standard classifier layer (e.g., MLP), treating intent labels as indices;
%These approaches leverage pretrained language models (PLMs) equipped with a standard classifier layer (e.g., MLP). Intent labels are regarded as indices.
2) \textbf{Example-based approaches} \cite{dnnc, example_driven, convfit}, which learn to compare the similarities between different examples and classify an input utterance based on the closest neighbor in the training data; 
3) \textbf{Intent semantic aware approaches} \cite{qu2021, incrementalfstc, Du2022LearningTS}, which explicitly incorporate intent label words during training.
However, both classifier-based and example-based methods disregard label semantic information, which is an important source of supervision in few-shot scenarios.
Exiting intent semantic aware approaches also suffer from different drawbacks, such as relying on large-scale pretraining datasets and only partially using intent labels. More details on related work are discussed in Appendix \ref{related_work}.
%they need an additional step to select the top-k intent  and it's not guaranteed that the ground-truth labeled is selected in the top-k intents. 
%\citet{qu2021} and \citet{incrementalfstc} cast ID as textual entailment (\standardentail), treating utterances and intents as premises and hypotheses, respectively. \citet{Du2022LearningTS} propose \contextTE~and \parallelTE, embedding the information of multiple intents into a single \standardentail~pair. \citet{LSAP} incorporate intent semantics into generative models via pretraining.
%Standard classifier-based and example-based methods ignore label semantic information, which is the important supervision source in zero- and few-shot learning.
%\textcolor{red}{However, these intent semantic aware methods leverage intent information indirectly by transferring intent detection into other task formats, such as textual entailment or text generation.
%(e.g., casting ID as \standardentail~or text generation). 
%The gap between different task formats may limit their performance.}
%Furthermore, most existing few-shot intent detection methods rely on extra pre-training data to obtain decent performance.
%extra training phase and data to enhance the model generalization capacity. 

To solve these issues, we propose an end-to-end intent semantic aware model, \parallelID. 
It concatenates each utterance with the entire intent label set as the input and then encodes them simultaneously. 
In this way, the semantic information of all intents is fully utilized and integrated with utterances.
The encoded embeddings of labels and utterances are subsequently used for contrastive learning.
We define a new contrastive learning paradigm by comparing the representations of utterances and intents directly.
This approach ensures utterances are moved closer to their gold intents while distancing them from any incorrect ones.
Furthermore, we introduce a novel pretraining strategy for \parallelID~that leverages indirect supervision from paraphrase identification datasets. Through this strategy, the model develops the ability to understand semantic similarities and distinctions among sentences, generalizing its comprehension to unseen intents in zero-shot intent detection tasks.

% During training, a positive pair is formed by pairing an utterance with its corresponding gold intent, while negative pairs are constructed by pairing an utterance with other incorrect intents.

% that uses contrastive learning to compare an utterance with the entire intent label set. 

%Different from \citet{Du2022LearningTS}, \parallelID~does not need to select the top-$k$ filtering model. It works on the entire intent space other than the top-$k$ intents. 
%\parallelID~ splits all intents into multiple groups and each training example of \parallelID~contains one utterance with one group of intents. With this, \parallelID~can perform a one-to-many comparison for each utterance. By constructing training examples for each utterance with different groups of intents, \parallelID~ is able to compare each utterance with all the intents.
%in a one-to-many comparison format.
%During training, \parallelID~ maps each utterance and intent to a semantic space with a PLM, and then perform contrastive learning between the utterance and intents in the same training example.
%This approach ensures that utterances are pushed closer to their gold intents and away from all the incorrect ones.

To demonstrate the effectiveness of our proposed model, we conduct experiments on three fine-grained intent detection tasks:  BANKING77 \cite{banking77}, HWU64 \cite{hwu64} and CLINC150 \cite{clinc150}, under low-shot settings (0-, 1-, 3- and 5-shot).
The results show that \parallelID~ is especially effective in extreme few-shot scenarios, with an average improvement of 4.62\% in 1-shot and 2.60\% in 3-shot settings over the state-of-the-art (SOTA) without any pretraining. 
Our model also achieves SOTA in 5-shot scenarios with pretraining on out-of-domain (OOD) data.
Furthermore, \parallelID~ shows great cross-domain generalization capabilities in the zero-shot setting when further pretrained on paraphrasing identification datasets.
%projects utterances and labels in parallel. 
%Different from \cite{Du2022LearningTS}, \parallelID~works on entire intents space instead of top-$k$ by dividing all intents into multiple groups. Each input pair of \parallelID~contains an utterance and a group of intents and then is mapped to a semantic space by a PLM. Further, \parallelID~ directly leverages intent semantics by performing contrastive learning between the utterance and intents in the same pair. In this way, utterances are pushed closer to their gold intents and away from the incorrect ones. Our work focuses on fine-grained intent detection tasks and is evaluated on three datasets:  BANKING77 \cite{banking77}, HWU64 \cite{hwu64} and CLINC150 \cite{clinc150}, under 1-, 3- and 5-shot settings. 

%\textcolor{red}{However, \parallelTE~\cite{Du2022LearningTS} is a two-stage model. It first selects a top-$k$ filtering model for each task to narrow the intent space and then selects the gold intents from the filtered top-$k$ intents, with a potential performance cap if the top-$k$ model misses gold intents. }

Our contributions can be summarized as follows. First, 
we proposed an end-to-end \parallelID~ system that enables the comparison of an input utterance with all label candidates via a newly defined contrastive learning paradigm. To our knowledge, this is the first work that can encode the entire label space while modeling the intent identification problem.
%we propose an end-to-end intent semantic aware model, \parallelID~, which leverages all the intent labels for each utterance via contrasitve learning.
%directly leveraging intents semantics via contrasitve learning.
Second, we go beyond few-shot intent detection and further achieve zero-shot cross-domain generalization with a novel pretraining stage of our model: pretraining on paraphrase identification datasets. This is the first work that effectively uses indirect supervision from paraphrasing to handle zero-shot intent identification tasks.

\vspace{-0.05in}
\section{Methods}
\vspace{-0.05in}
%In this paper, we propose an end-to-end semantic aware model, \parallelID~, which utilizes a different input construction strategy and performs contrastive learning between utterances and intents. Furthermore, we propose a pretraining strategy for \parallelID~ to make it benefit from out-of-domain (OOD) and paraphrase datasets.

%Compared to the two-stage \parallelTE, \parallelID~is an end-to-end model with a different input construction strategy and performs contrastive learning between utterances and intents. We also propose a pretraining strategy for \parallelID~ to make it benefit from out-of-domain (OOD) and paraphrase datasets.

% We also show that \parallelID~has a strong cross-domain generalization ability in few-shot and zero-shot scenarios and present a method to enhance \parallelID~ by pretraining on public datasets.

% 1. pair construction
% \vspace{-0.05in}
\subsection{\parallelID~ Input Sequence Construction}
\begin{figure}
 \setlength{\belowcaptionskip}{-10pt}
 \setlength{\abovecaptionskip}{5pt}
    \centering
    \includegraphics[width=0.95\linewidth]{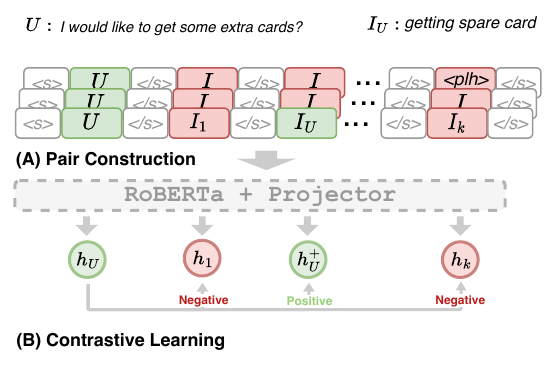}
    \caption{Overview of \parallelID. (A) shows the input sequence construction strategy. $I_U$ indicates the gold intent of $U$. (B) shows positive and negative instances in contrastive learning.}
    \label{fig_model}
\end{figure}

% \vspace{-0.05in}
\label{pair_construction}
Incorporating additional labels as contexts alongside utterances within an input sequence can help the model make a better decision \cite{Du2022LearningTS}. \parallelID~concatenates each utterance with the complete intent label set as the input and encodes them together.
However, the limitation of the maximum input sequence length makes it impractical to include all labels in a single sequence, especially when the label space is large.
Therefore, given an intent detection task with $n$ intents, we take every $k$ intents as a group and then all the intents will be divided into $m = \lceil n/k \rceil$ groups. To keep each group with a consistent number of elements, we introduce a special placeholder token \textit{<plh>}. For the group whose number of intents $(s)$ is less than $k$, we fill it with $(k-s)$ \textit{<plh>}s to maintain consistency.
Each utterance $U$ is then duplicated $m$ times and appended with the $m$ groups of intents, respectively, yielding $m$ input sequences, as shown in Figure~\ref{fig_model}(A). In this way, the model works on the entire intent space and incorporates multiple intents into a single sequence.
In practice, we choose $k$ by minimizing $n \bmod k$ to avoid introducing too many \textit{<plh>}s. 
Each input sequence is then shuffled $k$ times during training to perform data augmentation, which is beneficial for few-shot tasks. 
%It also prevents the model from overfitting on the gold intents positions in the input sequence.
% 2. contrastive learning
% add some more intro on CL or refs.

\subsection{Intent-Aware Contrastive Learning}
\label{contrastive_learning}
To better exploit intent semantic information, we apply contrastive learning (CL) between utterances and intents. 
% move to introduction?
Previous works perform CL between input texts and their augmented views \cite{simcse, consert, simclr, moudisentangled}, or instances under different classes \cite{cpft, sclforplm, detect}, while we directly perform CL between input texts (utterances) and classes (intents) since intents usually contain useful semantic meanings. Therefore, the representations of intents can be regarded as different views of utterances or explicit cluster centroids. 
As shown in Figure~\ref{fig_model} (B), we feed the input sequence constructed from Section~\ref{pair_construction} into RoBERTa and obtain a list of token-level representations. Then for the utterance $U$ and all the intents $I_{j}, j \in \{1, ..., k\}$ in the input sequence, we average the representations of their corresponding tokens as their representations $z_U$ and $z_j, j \in \{1, ..., k\}$. Finally, we use a shared MLP projector to project $z_U$ and $z_j$ into the same semantic space, yielding the final utterance representation $h_U$ and intent representation $h_j$. We use $h_U^+$ to denote the representation of the gold intent $I_U$ for $U$.
Let $sim(u, v)$ denote the cosine similarity between $u$ and $v$.
The contrastive learning loss is then defined as follows:
\begin{equation}
\label{cl_loss}
    l = -\frac{1}{N_b} \sum^{N_b}_{i=1}log\frac{e^{sim(\mathbf{h}_U^i, \mathbf{h}_U^{i+})/\tau}}{\sum^{k}_{j=1}e^{sim(\mathbf{h}_U^i, \mathbf{h}_j^i)/\tau}}
\end{equation}
, where $N_b$ is the batch size. $h_*^i$ means the representation in the $i$-th input in a batch.  $\tau$ is the temperature parameter.  Through the contrastive loss, \parallelID~ pushes utterances closer to their gold intents and meanwhile drives them away from all the incorrect intents. For the input sequences that do not include gold intents, we simply set the numerator of Equation~\ref{cl_loss} to 1 so the model only pushes the utterance away from other intents.

%Since we concatenate all labels, shuffling their order in the sequence can create a large number of training instances and different contrastive pairs. This is especially beneficial for both few-shot and contrastive learning. 
By shuffling the order of concatenated intents within the sequence, a large number of training instances and distinct contrastive pairs can be generated. This is particularly beneficial for both few-shot and contrastive learning.
Furthermore, a more robust model is yielded since we enforce the model to recognize the correct intent regardless of its location in the concatenated input sequence.

%\subsection{Pretraining on OOD and Paraphrase Data}
\subsection{Zero-shot Intent Detection via Paraphrase Identification Pretraining}

% % enhance parallel-ID with indirect supervision? 
% Although \parallelID~achieves a remarkable performance by only fine-tuning on the target task, pretraining on OOD data and public paraphrase datasets can \textit{further improve its performance} and \textit{enable its cross-domain zero-shot learning ability}.

% \paragraph{pretraining on OOD Data.} 
% For the few-shot setting, we pretrained the model on the OOD data to further improve the performance.
% Given an intent detection task, we treat other data which do not have domain overlapping with this task as OOD data.
% %We treat intent detection tasks without domain overalapping of the target task as OOD data. 
% During OOD pretraining, the intent space is a union of intents from all the OOD data. We follow the same sequence construction strategy and training process stated in Section~\ref{pair_construction} and Section~\ref{contrastive_learning}.

%\paragraph{Pretraining on Paraphrase Detection Data.} 
Paraphrase identification \cite{paraphrase} is a task that aims at identifying if two sentences have the same meaning.
Pretraining on the paraphrase detection identification dataset encourages models to capture the essence of utterances rather than relying solely on surface-level patterns. This can improve the ability of \parallelID~ to distinguish between similar intents that might have subtle differences in wording or phrasing.
Therefore, we propose a novel pretraining stage to effectively use indirect supervision from paraphrasing for zero-shot intent detection task.
Specifically, for each sentence, we regard its paraphrase as its gold label and select other $t$ most similar sentences with the help of Sentence-BERT \cite{sbert} as the negative labels. For the target task with $n$ intents, we set $t = n - 1$ so each sentence has totally $n$ labels to compare as well.
We set $k$, which is the number of intents in each group, the same as the target task to keep the pretraining and downstream tasks consistent.

Pretraining on out-of-domain (OOD) data can further improve the performance of \parallelID. 
Given an intent detection task, we treat other data which do not have domain overlapping with this task as the OOD data.
%We treat intent detection tasks without domain overalapping of the target task as OOD data. 
During OOD pretraining, the intent space is a union of intents from all the OOD data. We follow the same sequence construction strategy and training process stated in Section~\ref{pair_construction} and Section~\ref{contrastive_learning}.

%For both OOD and paraphrase pretraining, 

% 3. OOD cross-domain transfer
% table for data stats

\section{Experiments}
We conduct experiments on three intent detection tasks under both few-shot and zero-shot settings.
\paragraph{Datasets.} Our model is evaluated on three fine-grained intent detection datasets: BANKING77 \cite{banking77}, HWU64 \cite{hwu64}, and CLINC150 \cite{clinc150}. Each dataset contains one or multiple domains and a fine-grained intent space.
Dataset statistics are shown in Appendix Table \ref{tab_data_stat}. We randomly sample $10\%$ training data as the \textit{dev set}, following \citet{cpft} and \citet{dialoglue}.
For each target task, we use the other two datasets excluding the similar domains and intents as its OOD pretraining data. For example, we take BANKING77 as the target task, which contains banking-related intents. To obtain its OOD pretraining data, we combine the data from HWU64 and CLINC150 but remove the ``Banking'' and ``Credit Cards'' domains.
The statistics of the OOD data for each target task are shown in Appendix Table \ref{tab_ood_stat}. We also sample $10\%$ OOD data as the \textit{dev set} for the OOD pretraining.

For paraphrase detection pretraining, we leverage the public Quora Question Pairs (QQP) dataset\footnote{https://quoradata.quora.com/First-Quora-Dataset-Release-Question-Pairs}. 
To incorporate more sentences in a single pair, we filter short sentence pairs from QQP by setting the max number of words and characters in a sentence as 10 and 40, respectively. The filtered QQP dataset contains 31,412 paraphrase pairs.
%public Quora Question Pairs (QQP) \footnote{https://quoradata.quora.com/First-Quora-Dataset-Release-Question-Pairs}.

    \begin{table*}
     \setlength{\belowcaptionskip}{-10pt}
 \setlength{\abovecaptionskip}{5pt}
 \scalebox{0.845}{
    \centering
    
    \begin{tabular}{lccccccccc} 
    \toprule
    \multicolumn{1}{c}{\multirow{2}{*}{Model}} & \multicolumn{3}{c}{BANKING77} & \multicolumn{3}{c}{HWU64} & \multicolumn{3}{c}{CLINC150} \\ 
    \cmidrule(lr){2-4}  \cmidrule(lr){5-7}  \cmidrule(lr){8-10}
    \multicolumn{1}{c}{} & 1-shot & 3-shot & 5-shot & 1-shot & 3-shot & 5-shot  & 1-shot & 3-shot & 5-shot \\ 
    \midrule
    RoBERTa & 35.31\tiny{(2.22)} & 64.78\tiny{(0.76)}  & 75.47\tiny{(1.69)}  & 40.34\tiny{(2.66)}  & 67.44\tiny{(1.37)}  & 75.71\tiny{(1.26)}  & 48.48\tiny{(1.03)}  & 80.91\tiny{(2.10)}  & 86.82\tiny{(1.40)}  \\
    CPFT & 40.71\tiny{(2.25)}  & 71.57\tiny{(0.34)}  & 79.73\tiny{(0.52)}  & 50.61\tiny{(2.18)}  & 72.91\tiny{(2.32)}  & 79.82\tiny{(1.64)}  & 58.58\tiny{(0.57)}  & 83.12\tiny{(0.43)}  & \underline{90.60}\tiny{(0.70)}  \\
    RoBERTa-SPI & 43.81\tiny{(2.01)}  & 65.63\tiny{(0.66)}  & 72.20\tiny{(1.45)}  & 53.75\tiny{(2.13)}  & 70.95\tiny{(1.14)}  & 75.58\tiny{(1.24)}  & 67.57\tiny{(0.99)}  & 83.31\tiny{(1.78)}  & 87.76\tiny{(1.08)}  \\
    DNNC & 30.23\tiny{(2.51)}  & 72.21\tiny{(1.02)}  & \underline{79.94}\tiny{(1.77)}  & 29.77\tiny{(1.45)}  & 75.25\tiny{(2.69)}  & 79.31\tiny{(0.19)}  & 30.17\tiny{(2.33)}  & 87.07\tiny{(0.44)}  & 90.44\tiny{(1.03)}  \\
    Context-TE & 64.22\tiny{(0.50)}  & 73.27\tiny{(0.43)}  & 77.07\tiny{(0.57)}  & 64.39\tiny{(1.64)}  & 73.45\tiny{(1.72)}  & 78.16\tiny{(0.94)}  & 74.71\tiny{(0.91)}  & 84.56\tiny{(1.54)}  & 87.61\tiny{(0.79)}  \\
    Parallel-TE & 64.34\tiny{(1.23)}  & 72.20\tiny{(1.00)}  & 76.23\tiny{(0.37)}  & 61.96\tiny{(0.46)}  & 74.10\tiny{(0.51)}  & 78.10\tiny{(1.40)}  & 74.54\tiny{(0.81)}  & 83.66\tiny{(0.84)}  & 86.61\tiny{(0.58)}  \\ 
    \midrule
    \parallelID & \underline{\textbf{66.36}}\tiny{(0.46)}  & \underline{\textbf{76.13}}\tiny{(0.45)}  & 79.75\tiny{(0.78)}  & \underline{\textbf{68.77}}\tiny{(1.94)}  & \underline{\textbf{78.16}}\tiny{(2.04)}  & \underline{\textbf{79.89}}\tiny{(0.30)}  & \underline{\textbf{77.63}}\tiny{(0.63)}  & \underline{\textbf{87.09}}\tiny{(1.44)}  & 89.88\tiny{(0.81)}  \\ 
     ~~w/ OOD & \textbf{67.93}\tiny{(0.28)}  & \textbf{76.92}\tiny{(0.18)}  & 
     \textbf{80.51}\tiny{(0.88)}  & \textbf{73.17}\tiny{(0.37)}  & \textbf{79.95}\tiny{(0.68)}  & \textbf{82.50}\tiny{(1.05)}  & \textbf{79.21}\tiny{(0.43)}  & \textbf{88.01}\tiny{(1.46)}  & \textbf{90.76}\tiny{(0.63)}  \\ 
    \bottomrule 
    \end{tabular}
}
    
    \caption{Test accuracy (\%) and standard deviation on three dataset under three few-shot scenarios. The first and second highest results are formatted in \textbf{bold} and \underline{underline}, respectively.}
    \label{tab_few_shot}
    \end{table*}

\paragraph{Baselines.}
We compare our method with six baselines in the three categories as we described in Section \ref{intro}.
Standard classifier-based baselines:
1) RoBERTa with a standard classifier.
2) CPFT \cite{cpft} performs self-supervised CL on multiple intent detection datasets and applies supervised CL on the target task.  
3) RoBERTa-SPI \cite{spi} introduces two regularizers to improve supervised pretraining via isotropization. 
For example-based approaches, we consider its prior SOTA model:
4) DNNC \cite{dnnc}, which identifies intents by finding the nearest neighbors of utterances in the training set and is pretrained on three natural language inference datasets.
For intent semantic aware methods, we compare:
5) Context-TE and 6) Parallel-TE \cite{Du2022LearningTS}. These approaches incorporate multiple intents into a single textual entailment sequence. Context-TE relies on indirect supervision from MNLI~\cite{mnli}.
Parallel-TE is more comparable to \parallelID, it also encodes utterances and intents simultaneously, but it is a pipeline model that only selects top-$k$ intents for each utterance.
Among all the baselines, CPFT and DNNC are the prior SOTA models for 5- and 10-shot settings. Our paper considers more challenging scenarios, specifically the 1- and 3-shot settings.
The implementation of the baselines and our model are detailed in Appendix \ref{implementation}.

\paragraph{Results.} For few-shot experiments, we conduct three runs with distinct training data samples, following~\citet{Du2022LearningTS}.
%we sample the training examples three times and use them for all the methods. 
Table~\ref{tab_few_shot} shows the average accuracy and standard deviation on three datasets under 1-, 3-, and 5-shot settings.  \parallelID~outperforms all the baselines remarkably in 1- and 3-shot scenarios across three datasets without any pretraining. For example, \parallelID~ improves the state-of-the-art result for 1-shot on HWU64 by $6.81\%$. After pretraining on OOD data, the improvement percentage increases to 13.64\%.
For the 5-shot setting, \parallelID~ achieves comparable results without pretraining and outperforms all the baselines with pretraining on OOD.
The example-based model DNNC performs extremely poorly on 1-shot tasks even though it achieves good performances in 5-shot, showing its limitation when training resources are extremely scarce.
% Despite CPFT's utilization of CL, it only forms contrastive pairs among utterances while neglecting the intent semantics. For the 1- and 3-shot performance of 

\begin{figure}
 \setlength{\belowcaptionskip}{-10pt}
 \setlength{\abovecaptionskip}{5pt}
    \centering
    \includegraphics[width=\linewidth]{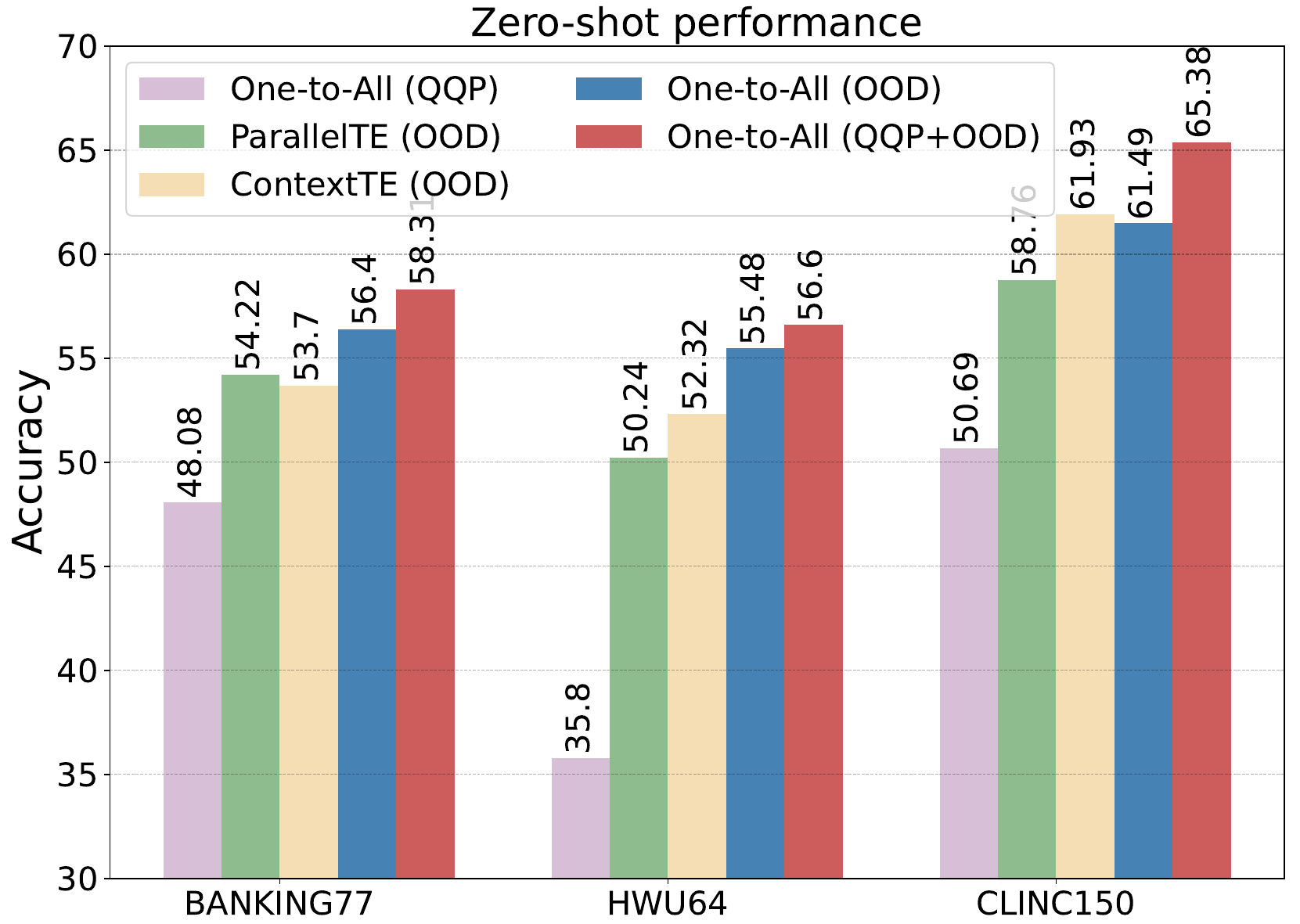}
    \caption{Zero-shot performance. OOD or QQP indicates that the model is pretrained on OOD or QQP data.}
    \label{fig_zero_shot}
\end{figure}

%To further investigate the cross-domain transfer ability of \parallelID, we conduct experiments under the zero-shot setting.
For the zero-shot setting, we first did preliminary experiments evaluating all baselines and our model on the target tasks without extra pretraining.
The results are reported in Appendix Table~\ref{tab_zero_shot}.
Despite poor performance across all models, \parallelID~remains the top-performing approach.
Then we pretrain the model on OOD/QQP data without accessing any in-domain training data. The results are shown in Figure~\ref{fig_zero_shot}. 
The zero-shot performance of \parallelID~ exhibits significant improvement following pretraining on QQP data, indicating the effectiveness of our novel pretraining strategy.
%evaluate it under the zero-shot setting, i.e., for a target task, the model is only trained on the OOD data and then directly tested.
Comparing the zero-shot results with the few-shot results in Table \ref{tab_few_shot}, we can observe that the performance of \parallelID~(OOD) under the zero-shot setting even outperforms some baselines (RoBERTa, CPFT, and DNNC) under the 1-shot setting.
Thus, we compare our model, \parallelID, with the two strongest baselines, Context-TE and Parallel-TE, in the zero-shot setting.
%The zero-shot \parallelID~ outperforms almost all the baselines under the 1-shot setting except RoBERTa-SPI(on CLINC150), ContextTE and ParallelTE.
%, while RoBERTa-SPI has already been trained on OOD data, as we stated in baselines.
As shown in Figure \ref{fig_zero_shot}, \parallelID~(OOD) outperforms both Context-TE (OOD) and Parallel-TE (OOD) in most settings. 
%With pretraining on both OOD and QQP, \parallelID~ is further boosted by first performing paraphrase detection pretraining and then OOD pretraining, outperforming both ContextTE and ParallelTE by a large margin.
%. except ContextTE on CLINC150, but ContextTE relies on the indirect supervision from MNLI. 
The zero-shot performance of \parallelID~ is further boosted by pretraining on both OOD and QQP data, outperforming both Context-TE and Parallel-TE by a large margin. 
Once again, this finding highlights the effectiveness of utilizing indirect supervision from paraphrase identification datasets.
%by first performing paraphrase detection pretraining and then OOD pretraining, outperforming both ContextTE and ParallelTE by a large margin.

\paragraph{Analysis} 

%We investigate why \parallelID~performs better than other intent semantic aware models, such as ParallelTE, 

We investigate how our end-to-end design impacts our model performance by comparing it with Parallel-TE through a case study on BANKING77. We run \parallelID~and Parallel-TE under the 3-shot setting three times.
The top-$k$ filtering step in Parallel-TE misses gold intents for 89 utterances in the \textit{test set}, while \parallelID~can correctly predict 34.7 of them on average, which brings a $1.1\%$ improvement given the size of \textit{test set} is 3080.
This suggests that our end-to-end design effectively identifies partial intents that may have been overlooked by the pipeline system.
% 2) We also find that \parallelID~can better understand intent semantics instead of simply performing word matching between utterances and intents. For example, given the utterance \textit{``Can I purchase extra non-virtual cards?''}, \parallelID~identifies its correct intent \textit{``getting spare card''} while ParallelTE predicts it as \textit{``getting virtual card''}.

\section{Conclusion}

In this paper, we propose an end-to-end intent semantic aware intent detection model \parallelID~ to fully leverage intent semantics via contrastive learning. Experiments show that it is especially effective when training resource is scarce.

%Experiments show that our method is especially effective when training resource is extremely scarce.

\section*{Limitations}

Although we perform OOD pretraining on our model and gain performance improvement, the OOD data we use in our experiment is only from at most two datasets. 
There are many publicly available intent detection datasets from different domains, such as ATIS \cite{atis} and SNIPS \cite{snips}, can be used for the OOD pretraining.
We believe pretraining on large-scale OOD datasets can further boost the  performance of our model, and we leave it for our future work.

\section*{Acknowledgments}
The authors appreciate the reviewers for their insightful comments and suggestions.
This work is supported in part by NSF under grant III-2106758.

% \usepackage{multirow}

\bibliography{anthology,custom}

\begin{thebibliography}{30}
\expandafter\ifx\csname natexlab\endcsname\relax\def\natexlab#1{#1}\fi

\bibitem[{Burnyshev et~al.(2021)Burnyshev, Bout, Malykh, and Piontkovskaya}]{infobert}
Pavel Burnyshev, Andrey Bout, Valentin Malykh, and Irina Piontkovskaya. 2021.
\newblock {I}n{F}o{BERT}: Zero-shot approach to natural language understanding using contextualized word embedding.
\newblock In \emph{Proceedings of RANLP}, pages 208--215.

\bibitem[{Casanueva et~al.(2020)Casanueva, Tem{\v{c}}inas, Gerz, Henderson, and Vuli{\'c}}]{banking77}
I{\~n}igo Casanueva, Tadas Tem{\v{c}}inas, Daniela Gerz, Matthew Henderson, and Ivan Vuli{\'c}. 2020.
\newblock Efficient intent detection with dual sentence encoders.
\newblock In \emph{Proceedings of the 2nd Workshop on Natural Language Processing for Conversational AI}, pages 38--45.

\bibitem[{Chen et~al.(2020)Chen, Kornblith, Norouzi, and Hinton}]{simclr}
Ting Chen, Simon Kornblith, Mohammad Norouzi, and Geoffrey Hinton. 2020.
\newblock A simple framework for contrastive learning of visual representations.
\newblock In \emph{Proceedings of ICML}, volume 119, pages 1597--1607.

\bibitem[{Coucke et~al.(2018)Coucke, Saade, Ball, Bluche, Caulier, Leroy, Doumouro, Gisselbrecht, Caltagirone, Lavril, Primet, and Dureau}]{snips}
Alice Coucke, Alaa Saade, Adrien Ball, Th{\'e}odore Bluche, Alexandre Caulier, David Leroy, Cl{\'e}ment Doumouro, Thibault Gisselbrecht, Francesco Caltagirone, Thibaut Lavril, Ma{\"e}l Primet, and Joseph Dureau. 2018.
\newblock Snips voice platform: an embedded spoken language understanding system for private-by-design voice interfaces.
\newblock \emph{ArXiv}, abs/1805.10190.

\bibitem[{Du et~al.(2022)Du, Yin, Xia, and Yu}]{Du2022LearningTS}
Jiangshu Du, Wenpeng Yin, Congying Xia, and Philip~S. Yu. 2022.
\newblock Learning to select from multiple options.
\newblock \emph{ArXiv}, abs/2212.00301.

\bibitem[{Gao et~al.(2021)Gao, Yao, and Chen}]{simcse}
Tianyu Gao, Xingcheng Yao, and Danqi Chen. 2021.
\newblock {S}im{CSE}: Simple contrastive learning of sentence embeddings.
\newblock In \emph{Proceedings of EMNLP}, pages 6894--6910.

\bibitem[{Gunel et~al.(2021)Gunel, Du, Conneau, and Stoyanov}]{sclforplm}
Beliz Gunel, Jingfei Du, Alexis Conneau, and Veselin Stoyanov. 2021.
\newblock Supervised contrastive learning for pre-trained language model fine-tuning.
\newblock In \emph{Proceedings of ICLR}.

\bibitem[{He et~al.(2022{\natexlab{a}})He, Dai, Hui, Yang, Cao, Dong, Huang, Si, and Li}]{space2}
Wanwei He, Yinpei Dai, Binyuan Hui, Min Yang, Zheng Cao, Jianbo Dong, Fei Huang, Luo Si, and Yongbin Li. 2022{\natexlab{a}}.
\newblock {SPACE}-2: Tree-structured semi-supervised contrastive pre-training for task-oriented dialog understanding.
\newblock In \emph{Proceedings of COLING}, pages 553--569.

\bibitem[{He et~al.(2022{\natexlab{b}})He, Dai, Yang, Sun, Huang, Si, and Li}]{space3}
Wanwei He, Yinpei Dai, Min Yang, Jian Sun, Fei Huang, Luo Si, and Yongbin Li. 2022{\natexlab{b}}.
\newblock Unified dialog model pre-training for task-oriented dialog understanding and generation.
\newblock In \emph{Proceedings of SIGIR}, page 187–200.

\bibitem[{Hemphill et~al.(1990)Hemphill, Godfrey, and Doddington}]{atis}
Charles~T. Hemphill, John~J. Godfrey, and George~R. Doddington. 1990.
\newblock The {ATIS} spoken language systems pilot corpus.
\newblock In \emph{Speech and Natural Language: Proceedings of a Workshop Held at Hidden Valley, {P}ennsylvania, June 24-27,1990}.

\bibitem[{Lamanov et~al.(2022)Lamanov, Burnyshev, Artemova, Malykh, Bout, and Piontkovskaya}]{template}
Dmitry Lamanov, Pavel Burnyshev, Katya Artemova, Valentin Malykh, Andrey Bout, and Irina Piontkovskaya. 2022.
\newblock Template-based approach to zero-shot intent recognition.
\newblock In \emph{Proceedings of INLG}, pages 15--28.

\bibitem[{Larson et~al.(2019)Larson, Mahendran, Peper, Clarke, Lee, Hill, Kummerfeld, Leach, Laurenzano, Tang, and Mars}]{clinc150}
Stefan Larson, Anish Mahendran, Joseph~J. Peper, Christopher Clarke, Andrew Lee, Parker Hill, Jonathan~K. Kummerfeld, Kevin Leach, Michael~A. Laurenzano, Lingjia Tang, and Jason Mars. 2019.
\newblock An evaluation dataset for intent classification and out-of-scope prediction.
\newblock In \emph{Proceedings of EMNLP-IJCNLP}, pages 1311--1316.

\bibitem[{Lin et~al.(2022)Lin, Ma, Chen, Yang, Cheng, and Guang}]{detect}
Hongzhan Lin, Jing Ma, Liangliang Chen, Zhiwei Yang, Mingfei Cheng, and Chen Guang. 2022.
\newblock Detect rumors in microblog posts for low-resource domains via adversarial contrastive learning.
\newblock In \emph{Findings of NAACL}, pages 2543--2556.

\bibitem[{Liu et~al.(2019{\natexlab{a}})Liu, Eshghi, Swietojanski, and Rieser}]{hwu64}
Xingkun Liu, Arash Eshghi, Pawel Swietojanski, and Verena Rieser. 2019{\natexlab{a}}.
\newblock Benchmarking natural language understanding services for building conversational agents.
\newblock \emph{ArXiv}, abs/1903.05566.

\bibitem[{Liu et~al.(2019{\natexlab{b}})Liu, Ott, Goyal, Du, Joshi, Chen, Levy, Lewis, Zettlemoyer, and Stoyanov}]{roberta}
Yinhan Liu, Myle Ott, Naman Goyal, Jingfei Du, Mandar Joshi, Danqi Chen, Omer Levy, Mike Lewis, Luke Zettlemoyer, and Veselin Stoyanov. 2019{\natexlab{b}}.
\newblock Roberta: A robustly optimized bert pretraining approach.
\newblock \emph{ArXiv}, abs/1907.11692.

\bibitem[{Mehri and Eric(2021)}]{example_driven}
Shikib Mehri and Mihail Eric. 2021.
\newblock Example-driven intent prediction with observers.
\newblock In \emph{Proceedings of NAACL-HLT}, pages 2979--2992.

\bibitem[{Mehri et~al.(2020)Mehri, Eric, and Hakkani{-}T{\"{u}}r}]{dialoglue}
Shikib Mehri, Mihail Eric, and Dilek Hakkani{-}T{\"{u}}r. 2020.
\newblock Dialoglue: {A} natural language understanding benchmark for task-oriented dialogue.
\newblock \emph{CoRR}, abs/2009.13570.

\bibitem[{Mou et~al.(2022)Mou, He, Wu, Zeng, Xu, Jiang, Wu, and Xu}]{moudisentangled}
Yutao Mou, Keqing He, Yanan Wu, Zhiyuan Zeng, Hong Xu, Huixing Jiang, Wei Wu, and Weiran Xu. 2022.
\newblock Disentangled knowledge transfer for {OOD} intent discovery with unified contrastive learning.
\newblock In \emph{Proceedings of ACL (Short Papers)}, pages 46--53.

\bibitem[{Mueller et~al.(2022)Mueller, Krone, Romeo, Mansour, Mansimov, Zhang, and Roth}]{LSAP}
Aaron Mueller, Jason Krone, Salvatore Romeo, Saab Mansour, Elman Mansimov, Yi~Zhang, and Dan Roth. 2022.
\newblock Label semantic aware pre-training for few-shot text classification.
\newblock In \emph{Proceedings of ACL}, pages 8318--8334.

\bibitem[{Qu et~al.(2021)Qu, Hashimoto, Liu, Xiong, and Zhou}]{qu2021}
Jin Qu, Kazuma Hashimoto, Wenhao Liu, Caiming Xiong, and Yingbo Zhou. 2021.
\newblock Few-shot intent classification by gauging entailment relationship between utterance and semantic label.
\newblock In \emph{Proceedings of the 3rd Workshop on Natural Language Processing for Conversational AI}.

\bibitem[{Reimers and Gurevych(2019)}]{sbert}
Nils Reimers and Iryna Gurevych. 2019.
\newblock Sentence-{BERT}: Sentence embeddings using {S}iamese {BERT}-networks.
\newblock In \emph{Proceedings of EMNLP-IJCNLP}, pages 3982--3992.

\bibitem[{Socher et~al.(2011)Socher, Huang, Pennin, Manning, and Ng}]{paraphrase}
Richard Socher, Eric Huang, Jeffrey Pennin, Christopher~D Manning, and Andrew Ng. 2011.
\newblock Dynamic pooling and unfolding recursive autoencoders for paraphrase detection.
\newblock In \emph{Proceedings of {NeurIPs}}, pages 801--809.

\bibitem[{Vuli{\'c} et~al.(2021)Vuli{\'c}, Su, Coope, Gerz, Budzianowski, Casanueva, Mrk{\v{s}}i{\'c}, and Wen}]{convfit}
Ivan Vuli{\'c}, Pei-Hao Su, Samuel Coope, Daniela Gerz, Pawe{\l} Budzianowski, I{\~n}igo Casanueva, Nikola Mrk{\v{s}}i{\'c}, and Tsung-Hsien Wen. 2021.
\newblock {ConvFiT:} {C}onversational fine-tuning of pretrained language models.
\newblock In \emph{Proceedings of {EMNLP}}, pages 1151--1168. Association for Computational Linguistics.

\bibitem[{Williams et~al.(2018)Williams, Nangia, and Bowman}]{mnli}
Adina Williams, Nikita Nangia, and Samuel Bowman. 2018.
\newblock A broad-coverage challenge corpus for sentence understanding through inference.
\newblock In \emph{Proceedings of {NAACL-HLT}}, pages 1112--1122.

\bibitem[{Wu et~al.(2020)Wu, Hoi, Socher, and Xiong}]{tod}
Chien-Sheng Wu, Steven~C.H. Hoi, Richard Socher, and Caiming Xiong. 2020.
\newblock {TOD}-{BERT}: Pre-trained natural language understanding for task-oriented dialogue.
\newblock In \emph{Proceedings of EMNLP}, pages 917--929.

\bibitem[{Xia et~al.(2021)Xia, Yin, Feng, and Yu}]{incrementalfstc}
Congying Xia, Wenpeng Yin, Yihao Feng, and Philip Yu. 2021.
\newblock Incremental few-shot text classification with multi-round new classes: Formulation, dataset and system.
\newblock In \emph{Proceedings of NAACL-HLT}, pages 1351--1360.

\bibitem[{Yan et~al.(2021)Yan, Li, Wang, Zhang, Wu, and Xu}]{consert}
Yuanmeng Yan, Rumei Li, Sirui Wang, Fuzheng Zhang, Wei Wu, and Weiran Xu. 2021.
\newblock {C}on{SERT}: A contrastive framework for self-supervised sentence representation transfer.
\newblock In \emph{Proceedings of ACL}, pages 5065--5075.

\bibitem[{Zhang et~al.(2022)Zhang, Liang, Zhang, Zhan, Wu, Lu, and Lam}]{spi}
Haode Zhang, Haowen Liang, Yuwei Zhang, Li-Ming Zhan, Xiao-Ming Wu, Xiaolei Lu, and Albert Lam. 2022.
\newblock Fine-tuning pre-trained language models for few-shot intent detection: Supervised pre-training and isotropization.
\newblock In \emph{Proceedings of NAACL-HLT}, pages 532--542.

\bibitem[{Zhang et~al.(2021)Zhang, Bui, Yoon, Chen, Liu, Xia, Tran, Chang, and Yu}]{cpft}
Jianguo Zhang, Trung Bui, Seunghyun Yoon, Xiang Chen, Zhiwei Liu, Congying Xia, Quan~Hung Tran, Walter Chang, and Philip Yu. 2021.
\newblock Few-shot intent detection via contrastive pre-training and fine-tuning.
\newblock In \emph{Proceedings of EMNLP}, pages 1906--1912.

\bibitem[{Zhang et~al.(2020)Zhang, Hashimoto, Liu, Wu, Wan, Yu, Socher, and Xiong}]{dnnc}
Jianguo Zhang, Kazuma Hashimoto, Wenhao Liu, Chien-Sheng Wu, Yao Wan, Philip Yu, Richard Socher, and Caiming Xiong. 2020.
\newblock Discriminative nearest neighbor few-shot intent detection by transferring natural language inference.
\newblock In \emph{Proceedings of EMNLP}, pages 5064--5082.

\end{thebibliography}
\bibliographystyle{acl_natbib}

%\clearpage

\appendix

\section{Related Work}
\label{related_work}
Recent research on low-shot intent detection can be broadly classified into three main categories: standard classifier-based, example-based, and intent semantic aware approaches.

Standard-classifier methods discard intent semantics and usually require an extra pretraining process on the extra corpus.
For example, \citet{space2, space3, tod} pretrain models on large-scale dialog datasets.
\citet{cpft} pretrain a RoBERTa model on various intent datasets via self-supervised contrastive learning.
\citet{spi} propose two regularizers to improve its supervised pretraining step via isotropization.

Example-based approaches aim to learn the similarities between various examples and classify an input utterance by identifying its closest neighbor in the training data.
For instance, \citet{dnnc, example_driven} determines the intent of an utterance by searching for its nearest neighbors among all training utterances. These approaches also ignore the semantic information of intents.

Existing intent semantic aware approaches also exhibit various drawbacks and limitations.
For example, LSAP \cite{LSAP} incorporates intent semantics into generative models via pretraining.
More specifically, during the pretraining stage, LSAP takes partially masked utterance-intent pairs as the input and predicts the masked contents.
However, this approach relies on large-scale pretraining data to obtain decent performance.
\citet{qu2021} and \citet{incrementalfstc} cast ID as textual entailment (\standardentail), treating utterances and intents as premises and hypotheses, respectively. But these two models are only able to compare one single utterance with one single intent, which makes them unaware of other intent options. \citet{Du2022LearningTS} learn to select the best intent for an utterance by providing the top-$k$ intents for that utterance in one training example.
%learn to select the best intent by presenting one utterance with top k intents in one training example.
%Given an input utterance, these methods classify the input sentence based on the nearest neighbor examples in the training data. 
%Both classifier-based approaches and example-based methods ignore the label semantic information, which is an important supervision source for few-shot intent detection.
%These methods implicitly regard intents as latent cluster centroids, aiming to find the nearest neighbor examples from the training data given an input utterance and then make predictions according to the intent of found examples. 
%Standard classifier-based and example-based methods ignore label semantic information, which is the important supervision source in zero- and few-shot learning.
Despite providing a one-to-many comparison, \citet{Du2022LearningTS} are only able to view the top-$k$ intents rather than the entire intent label set during training.
Additionally, \citet{Du2022LearningTS} propose a pipeline model. Their performance is constrained by the accuracy of the first stage of the pipeline, which is identifying the top-$k$ intents.
Moreover, \citet{template} propose a template-based approach for modeling intents and utterances as sentence pairs. \citet{infobert} use a deep contextualized model to embed utterances and the natural language descriptions of user intents in zero-shot scenarios.

Different from all the literature we discussed above, we propose an end-to-end intent semantic aware system. By fully utilizing label semantics via contrastive learning, our model achieves SOTA performance even without pretraining on additional datasets.

\section{Dataset Statistics}

% \begin{table*}
% \centering
% \begin{tabular}{lcccccc} 
% \hline
%  & \multicolumn{3}{c}{Original} & \multicolumn{3}{c}{OOD} \\ 
% \hline
%  & \#domain & \#utterance & \#intent & \#domain & \#utterance & \#intent \\ 
% \hline
% BANKING77 & 1 & 13083 & 77 & 29 & 28030 & 183 \\
% HWU64 & 21 & 10030 & 64 & 11 & 35453 & 225 \\
% CLINC150 & 10 & 22500 & 150 & 21 & 9854 & 63 \\
% \hline
% \end{tabular}
% \end{table*}

\begin{table}[ht]
 \setlength{\belowcaptionskip}{-10pt}
 \setlength{\abovecaptionskip}{5pt}
\centering
\begin{tabular}{lccc} 
\hline
 Dataset & Domain &  Utterance & Intent  \\
\hline
BANKING77 & 1 & 13,083 & 77  \\
HWU64 & 21 & 10,030 & 64  \\
CLINC150 & 10 & 22,500 & 150\\
\hline
\end{tabular}
\caption{Dataset statistics.}
\label{tab_data_stat}
\end{table}
\begin{table}[ht]
 \setlength{\belowcaptionskip}{-10pt}
 \setlength{\abovecaptionskip}{5pt}
\centering
\begin{tabular}{lccc} 
\hline
 Target & Domain &  Utterance & Intent  \\
\hline
BANKING77 & 29 & 28,030 & 183  \\
HWU64 & 11 & 35,453 & 225  \\
CLINC150 & 21 & 9,854 & 63\\
\hline
\end{tabular}
\caption{Statistics of the OOD pretraining data used for each target task.}
\label{tab_ood_stat}
\end{table}

\section{Implementation Details.} 
\label{implementation}
All the baselines and \parallelID~ adopt RoBERTa-base \cite{roberta} as the backbones for a fair comparison.
\paragraph{Baseline implementation.} 
For the RoBERTa model, we implement it with the Hugging Face\footnote{https://huggingface.co/} library.
For RoBERTa-SPI, DNNC, Context-TE, and Parallel-TE, we directly run their open-source code under our experiment settings.
It is important to note that the original Context-TE and Parallel-TE models utilize RoBERTa-large as their base models. However, in our implementation, we replace it with RoBERTa-base to ensure a fair comparison.
Regarding CPFT, we strive to replicate its methodology to the best of our ability, given the unavailability of its code and pretrained checkpoints to the public. The original paper trained CPFT on a fixed set of 5-shot data, while in our experiments, we conduct three runs with uniquely sampled few-shot training data, which mitigates the potential influence of data sampling bias.
%Originally, RoBERT-SPI is pretrained on one intent detection dataset and then tested on others. In our implementation, we first pretrain it on the OOD datasets described in the Datasets section and then evaluate it on the target task.

\paragraph{\parallelID~implementation.} For few-shot tasks and OOD pretraining, we train the model for 10 and 3 epochs, respectively, and keep the best ones on the \textit{dev set}.
For paraphrase detection pretraining, we train the model for 3 epochs. 
% Other hyperparameters are listed in Appendix \ref{hyper}. 
All the training batch size is set to 8 and the learning rate is 2e-5. The output dimension of the MLP projector is set to 768 and the temperature parameter $\tau=0.1$. We set $k$ to 26, 32, 30 for BANKING77, HWU64, and CLINC150, respectively, according to the observations on their \textit{dev sets}. Details are discussed in Appendix \ref{appendix_analysis}.  

% \section{Hyperparameter Configuration}
% \label{hyper}
% All the training batch size is set to 8 and the learning rate is 2e-5. The output dimension of the MLP projector is set to 768 and the temperature parameter $\tau=0.1$. We set $k$ to 26, 32, 30 for BANKING77, HWU64, and CLINC150, respectively, according to the observations on their \textit{dev sets}. Details are discussed in Appendix \ref{appendix_analysis}.  

\begin{figure}[t]
\centering
\subfigure[HWU64]{\includegraphics[width=3.8cm]  {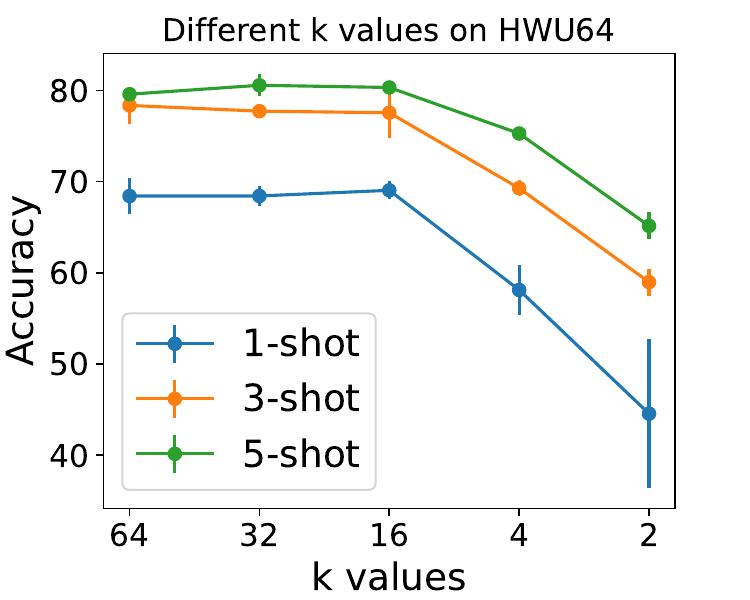}} \vspace{-0.05in}
\subfigure[BANKING77]{\includegraphics[width=3.8cm]{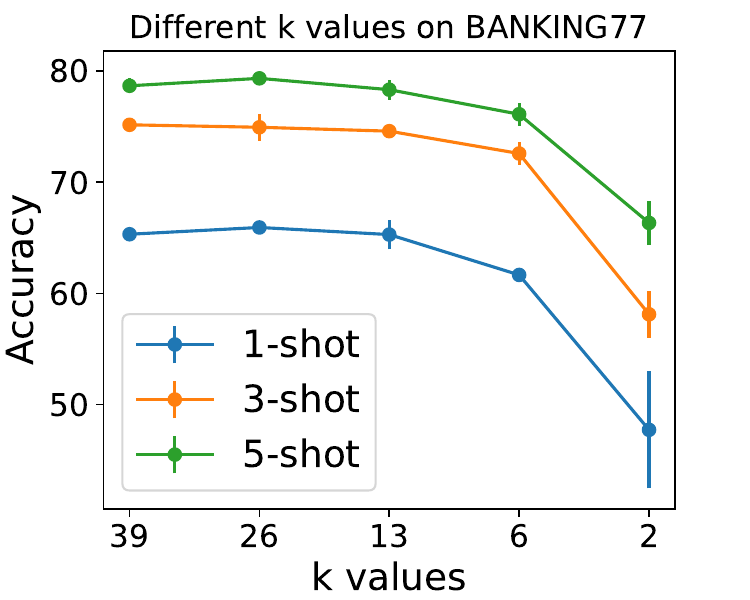}} \vspace{-0.05in}
\caption{Model performance with different $k$ values on the \textit{dev set} of HWU64 and BANKING77.}
\label{fig_k_influence}
\vspace{-0.05in}
\end{figure}
\section{Zero-shot Preliminary Experiments}

\begin{table}
\centering

\begin{tabular}{lccc} 
\toprule
\multicolumn{1}{c}{Model} & BANKING & HWU & CLINC \\ 
\cmidrule{1-1}\cmidrule(l){2-4}
RoBERTa & 1.65 & 1.77 & 2.01 \\
CPFT & 1.91 & 2.03 & 2.78 \\
RoBERTa-SPI & 1.90 & 1.87 & 2.52 \\
Context-TE & 1.04 & 1.20 & 1.43 \\
Parallel-TE & 1.66 & 2.14 & 2.24 \\ 
\cmidrule{1-1}\cmidrule(l){2-4}
 \parallelID & \textbf{6.04} & \textbf{13.05} & \textbf{5.84} \\
\bottomrule
\end{tabular}
\caption{Zero-shot performance (\%) without pretraining on additional datasets. DNNC is not included since it requires at least one training example.}
\label{tab_zero_shot}
\end{table}

\section{Additional Analysis}
\label{appendix_analysis}
We try to explore how the number of intents in each input sequence, $k$, influences the performance. We explore the influence of $k$ by conducting experiments on the \textit{dev set} of HWU64 and BANKING77 with different $k$ values. As shown in Figure~\ref{fig_k_influence}, the model performance stays similar when $k > 20$ but drops sharply when $k$ is below 10. 
This is probably due to the reduction of contrastive instances in a single pair.
Therefore, we set $k$ to 26, 32, 30 for BANKING77, HWU64, and CLINC150, respectively, and it can bring two benefits: 1) it is more friendly for the paraphrase detection pretraining as it is the max number of the sentences that can be incorporated into a single pair; 2) it minimizes $n \bmod k$ as we discussed in Section~\ref{pair_construction}.

\end{document}